\documentclass[10pt,twocolumn,letterpaper]{article}

\usepackage{wacv}              

\usepackage{times}    
\usepackage{xspace}
\usepackage[dvipsnames]{xcolor}
\usepackage{graphicx}
\usepackage{amsmath}
\usepackage{amssymb}
\usepackage{booktabs}
\usepackage{multirow}
\usepackage{float}
\usepackage[ruled,vlined]{algorithm2e}
\usepackage[numbers,sort&compress]{natbib}

\usepackage[pagebackref,breaklinks,colorlinks]{hyperref}

\usepackage[capitalize]{cleveref}
\crefname{section}{Sec.}{Secs.}
\Crefname{section}{Section}{Sections}
\Crefname{table}{Table}{Tables}
\crefname{table}{Tab.}{Tabs.}

\def\wacvPaperID{2760} 
\def\confName{WACV}
\def\confYear{2025}

\begin{document}


\title{Only What’s Necessary: Pareto-Optimal Data Minimization for Privacy Preserving Video Anomaly Detection}

\author{
Nazia Aslam$^{1}$ \hspace{0.2cm}
Abhisek Ray$^{2}$ \hspace{0.2cm}
Thomas B. Moeslund$^{1}$ \hspace{0.2cm}
Kamal Nasrollahi$^{1,3}$ \\
$^{1}$Aalborg University, Denmark \hspace{0.2cm}
$^{2}$Aarhus University, Denmark \hspace{0.2cm}
$^{3}$Milestone System, Denmark \\
{\tt\small \{naas,tbm,kn\}@create.aau.dk, ar@ece.au.dk}
}

\maketitle

\begin{abstract}

    Video anomaly detection (VAD) systems are increasingly deployed in safety‑critical environments and require a large amount of data for accurate detection. However, such data may contain personally identifiable information (PII), including facial cues and sensitive demographic attributes, creating compliance challenges under the EU General Data Protection Regulation (GDPR). In particular, GDPR requires that personal data be limited to what is strictly necessary for a specified processing purpose. To address this, we introduce \textit{Only What’s Necessary}, a privacy-by-design framework for VAD that explicitly controls the amount and type of visual information exposed to the detection pipeline. The framework combines breadth-based and depth-based data minimization mechanisms to suppress PII while preserving cues relevant to anomaly detection. We evaluate a range of minimization configurations by feeding the minimized videos to both a VAD model and a privacy inference model. We employ two ranking-based methods, along with Pareto analysis, to characterize the resulting trade-off between privacy and utility. From the non-dominated frontier, we identify \textit{sweet spot} operating points that minimize personal data exposure with limited degradation in detection performance. Extensive experiments on publicly available datasets demonstrate the effectiveness of the proposed framework. Find the code \href{https://github.com/Rabusi/only-what-s-necessary/tree/main}{HERE}.

\end{abstract}


\section{Introduction}
\label{sec:intro}

Data minimization is a core principle under Article 5(1)(c) of the European Union’s General Data Protection Regulation (GDPR), which states that personal data must be adequate, relevant, and limited to what is necessary for the stated purpose of processing \cite{gdprinfo_art5}. It serves both as a legal requirement and a privacy-by-design guideline, requiring organizations to collect and process only the data needed to achieve clearly defined objectives. However, this requirement becomes increasingly challenging in large-scale analytics and machine learning pipelines, where the pursuit of performance improvements often leads to an increase in the volume, detail, and retention of training and evaluation data. Consequently, privacy risks escalate, not only due to direct exposure, such as unauthorized access or distribution, but also through indirect leakage, which can occur via inference attacks and model exploitation \cite{nasr2019comprehensive, thomas2022framework}. In response to these risks, multiple regulatory frameworks, including the EU’s GDPR \cite{gdprinfo_art5}, California’s CPRA \cite{US-Colorado-CPA}, Brazil’s LGPD \cite{BR-LGPD}, and others \cite{UK-GDPR-ICO, CN-PIPL, IN-DPDP-DM}, have established data minimization standards that limit the collection and use of personal data to what is strictly necessary for the intended purpose.

International data protection regulations emphasize the principle that personal data collection must be \textit{adequate}, \textit{relevant}, and \textit{strictly necessary} for explicitly defined purposes. This principle is grounded in the observation that excessive data collection rarely contributes proportionately to its intended utility, while consistently increasing the risks of abuse and accidental disclosure \cite{shanmugam2022learning, goldsteen2022data, sorscher2022beyond, paul2021deep}. As such, data minimization should be viewed not only as a compliance requirement but also as a design guideline aimed at reducing privacy and security risks in data-driven systems. Despite its regulatory importance, data minimization still lacks a clear operational definition, especially one grounded in a mathematical formulation. This becomes a problem in machine learning pipelines, where it is often unclear how to turn necessary into measurable requirements. This gap has led to recent attempts to formalize data minimization as an explicit, measurable objective in machine learning.

Recent work on data minimization has increasingly focused on performance-based formulations, where data collection is aligned with the needs of downstream tasks, and privacy is often treated as a secondary concern rather than an explicit, competing objective \cite{finck2021reviving, biega2020operationalizing}. This perspective can help preserve utility, but it typically does not quantify privacy leakage or offer a way to control it when privacy and performance move in opposite directions. In contrast, Ganesh \etal~\cite{ganesh2025data} emphasizes that data minimization is inherently context-sensitive and privacy-centric, not simply a generic data-reduction heuristic. They model the data minimization and utility performance as a bi-level optimization problem, which operationalizes legal minimization requirements within machine learning pipelines. The results reflect that reducing data volume does not guarantee privacy gains, motivating minimization strategies that explicitly quantify and manage privacy risk rather than assuming it arises automatically from reduction. Building on these insights, our work explores how far video data can be minimized while still maintaining acceptable anomaly detection performance and simultaneously limiting privacy leakage. Our central idea is to treat minimization as a controllable design mechanism and to evaluate it against two concrete outcomes: (i) detection utility and (ii) privacy leakage. To this end, we apply minimization along two complementary dimensions: breadth-based reduction (how much visual information is exposed) and depth-based reduction (how much sensitive detail remains). For each minimization configuration, we evaluate utility using a VAD model and quantify privacy leakage using a dedicated privacy inference model. This yields a set of privacy–utility operating points, which we analyze using Pareto frontier to identify non-dominated “sweet spot” that reduce personal data exposure with minimal loss in detection performance. In summary, our contributions are as follows:

\begin{enumerate}

    \item We define video data minimization as a privacy-utility trade-off problem, aiming to preserve anomaly detection performance while reducing privacy leakage.
    
    \item We develop a two-dimensional minimization framework consisting of depth-based and breadth-based reduction. 

    \item We propose three operating-point selection strategies: distance-based, weighted aggregation-based, and constraint-based to identify Pareto-optimal configurations and determine practical ``sweet spot'' settings for real-world deployment.
    
    \item We comprehensively evaluate minimization settings across both dimensions to construct a utility--privacy Pareto frontier on the publically available dataset.

\end{enumerate}


\section{Related Work}
\label{sec:related work}

Major data protection regulations converge on three key principles of data minimization to safeguard privacy: (i) Adequacy: The data processed must be sufficient in quality and quantity to achieve the stated purpose, without undermining the functionality of the system; (ii) Relevance: Each category of data processed must have a demonstrable and legitimate connection to the stated purpose. Data that does not provide meaningful value should neither be collected nor utilized (iii) Limitation/Necessity: The processing must be restricted to the minimum personal data required, both in terms of scope (which types of attributes, modalities, and views are included) and volume (how many samples, frames, cameras, or the granularity and density of the captured signal)

Together, these principles create a clear normative goal: collect and process only what is necessary. However, translating these principles into machine learning practice is challenging. Legal texts often leave room for interpretation, and machine learning pipelines typically lack direct mechanisms for verifying adequacy, relevance, or necessity in measurable terms. As a result, operationalizing data minimization presents several challenges, including ambiguities in legal frameworks, the absence of widely accepted technical definitions, a lack of standardized mathematical formulations, and limited implementation guidance for organizations \cite{finck2021reviving}. Despite these challenges, interest in data minimization within machine learning is growing.

Recent work on data minimization in machine learning have been classified into breadth-based and depth-based approaches. Breadth-based minimization limits what information is retained, for example, by restricting the set of features (feature selection) or coarsening them through feature generalization \cite{rastegarpanah2021auditing, goldsteen2022data, staab2024principle}. In contrast, Depth-based minimization limits how much data is used, typically by reducing the number of training records through subsampling or pruning \cite{paul2021deep, sorscher2022beyond}, or by stopping data collection once performance gains saturate, via performance-based criteria informed by scaling laws \cite{shanmugam2022learning}. Other work studies personalized minimization in recommended systems \cite{biega2020operationalizing, chen2023studying, niu2023leveraging}, but these settings are less directly transferable to general-purpose ML pipelines. Several studies also formalize data minimization as an optimization problem \cite{biega2020operationalizing, niu2023leveraging}. For example, Ganesh et al.~\cite{ganesh2025data} model minimization as a bi-level optimization task, where the outer problem controls minimization and the inner problem evaluates downstream performance under a specified threat model. Despite these advances, much of the literature emphasizes dataset reduction \cite{biega2020operationalizing, chen2023studying, niu2023leveraging, rastegarpanah2021auditing, shanmugam2022learning}, often with limited attention to measurable privacy risk \cite{leemann2024prefer}. While some studies consider related notions such as information loss \cite{goldsteen2022data}, they do not directly quantify real-world privacy leakage. In contrast, our work makes privacy leakage a first-class criterion alongside task utility when assessing data minimization. We employed breadth and depth-based minimization settings and evaluate each configuration with two complementary signals: anomaly detection performance from a VAD model and privacy leakage measured by a privacy inference model. The resulting set of privacy–utility operating points enables Pareto analysis and a principled selection of deployable sweet spot configurations that reduce data exposure while limiting privacy risk.


\section{Method}
\label{method}

\begin{figure*}[h]
    \centering
    \begin{minipage}[t]{0.7\textwidth}
        \centering
        \includegraphics[height=6.2cm, width=12.7cm]{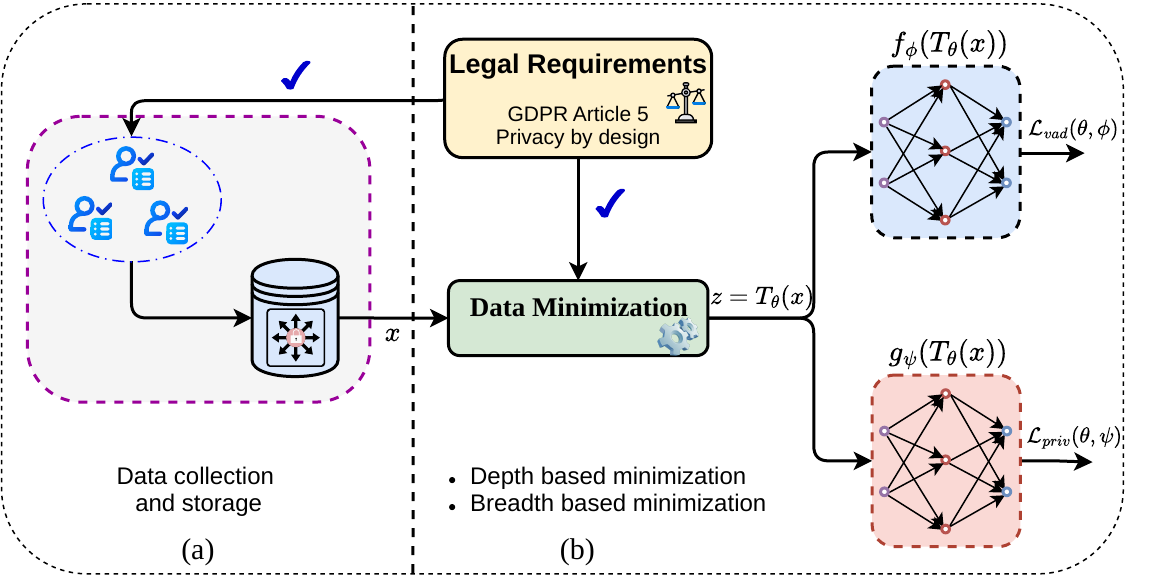}
    \end{minipage}\hfill
    \begin{minipage}[t]{0.3\textwidth}
        \centering
        \includegraphics[height=6.4cm, width=5cm]{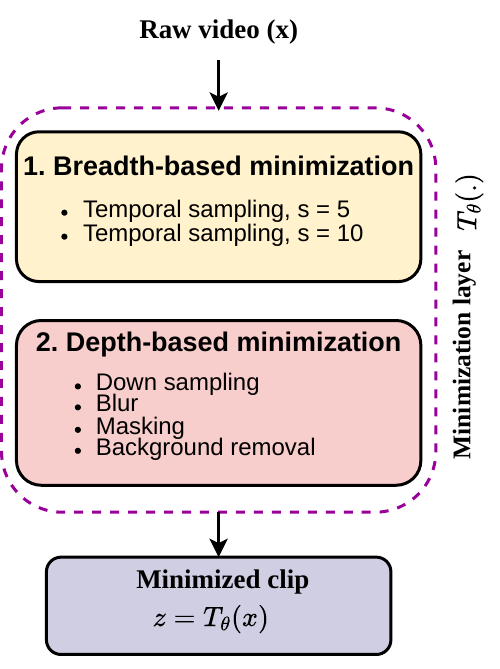}
    \end{minipage}
    \caption{Overview of the proposed data minimization framework. Given an input video clip $x \in \mathbb{R}^{T \times H \times W \times C}$, a configurable minimization module $T_{\theta}$ produces a reduced representation $z=T_{\theta}(x)$. The minimized video is then evaluated along two downstream objectives: anomaly detection using a VAD model $f_{\phi}$ and privacy leakage using a privacy inference model $g_{\psi}$.}
    \label{fig:architecture}
\end{figure*}

In many real-world settings, large video datasets were collected without strict data minimization constraints, creating potential privacy risks. We therefore redesign video anomaly detection under a data minimization setting, retaining only the necessary information. Figure~\ref{fig:architecture} illustrates the proposed framework for data minimization. Our method comprises three components: a configurable minimization module that transforms the input video into a reduced representation, downstream models that quantify anomaly detection utility and privacy leakage, and a Pareto-based selection procedure that identifies favorable operating points across minimization configurations. This formulation enables a systematic analysis of the utility-privacy trade-off induced by breadth-based and depth-based minimization strategies, yielding configurations that preserve strong anomaly detection performance while reducing privacy exposure.

\subsection{Data Minimization Framework}
\label{sec:framework}

Let $x \in \mathbb{R}^{T \times H \times W \times C}$ denote an input video clip, where $T$, $H$, $W$, and $C$ represent the number of frames, height, width, and channels, respectively. We define a configurable data minimization function $T_{\theta}$, parameterized by $\theta$, which transforms the raw video into a task-specific minimized representation

\begin{equation}
    z = T_{\theta}(x).
\end{equation}

The representation $z$ is subsequently used for two downstream evaluations: anomaly detection and privacy inference. In this way, the minimization module acts as the central control mechanism that determines how much visual information is preserved for the utility task and how much private information remains exposed.

\subsection{Minimization Strategy}
\label{sec:min_strategy}

To reduce privacy-relevant information in video data, we employ a combination of breadth-based and depth-based minimization strategies applied at complementary stages of the processing pipeline. Breadth-based minimization reduces the amount of data processed across time, while depth-based minimization reduces the information content within each selected frame.

\subsubsection{Breadth-based minimization}

Breadth-based minimization limits the temporal extent of the input by reducing how many frames or temporal segments are retained. In our framework, this is implemented through temporal sampling. Given a starting index $t_0$ and a fixed stride $s$, the selected frame indices are defined as:
\begin{equation}
    \mathcal{I}_{\text{clip}}(s,t_0)=\{\,t_0 + ns \mid n\in\mathbb{Z}_{\ge 0},\; t_0+ns \le T \,\},
    \label{eq:det_subsample}
\end{equation}
where $\mathcal{I}_{\text{clip}}(s,t_0)$ denotes the sampled temporal positions and $T$ is the total number of frames in the video. The resulting clip is

\begin{equation}
    V_{\text{clip}}=\{x_t\}_{t \in \mathcal{I}_{\text{clip}}}.
\end{equation}
By increasing the stride $s$, the model observes fewer temporal samples, thereby reducing the amount of raw visual information exposed during inference.

\subsubsection{Depth-based minimization}

Depth-based minimization reduces the amount of identifiable or context-rich information contained in each retained frame. In our framework, this is realized through several complementary operations. Background removal suppresses static scene context, masking restricts visibility to selected regions, and blurring or pixelation attenuates fine-grained appearance cues such as facial characteristics and texture details. We further apply spatial downsampling to remove high-frequency content and reduce overall visual fidelity. Collectively, these operations encourage the model to rely on coarse motion and structural patterns rather than identity-sensitive appearance information.

\subsection{Downstream Objectives}
\label{sec:objectives}

Each minimization configuration is evaluated using two downstream objectives: anomaly detection utility and privacy leakage.

\subsubsection{Anomaly detection objective}

Let $y_i \in \{0,1\}$ denote the ground-truth anomaly label associated with input video $x_i$. Given a minimized representation $T_{\theta}(x_i)$, the anomaly detection model $f_{\phi}$, parameterized by $\phi$, predicts the anomaly score. The corresponding objective is defined as:

\begin{equation}
    \mathcal{L}_{\mathrm{vad}}(\phi;\theta)
    =
    \frac{1}{N}\sum_{i=1}^{N}
    \ell_{\mathrm{vad}}\!\left(
    f_{\phi}\!\left(T_{\theta}(x_i)\right),\, y_i
    \right),
\end{equation}
where $\ell_{\mathrm{vad}}$ denotes the anomaly detection loss. This objective quantifies how well the minimized representation supports the downstream VAD task.

\subsubsection{Privacy leakage objective}

To assess residual privacy risk, we introduce a privacy inference model $g_{\psi}$, parameterized by $\psi$, which attempts to predict a private attribute $p$ from the same minimized representation:

\begin{equation}
    \hat{p} = g_{\psi}(z)=g_{\psi}\!\left(T_{\theta}(x)\right).
\end{equation}
For a dataset $\{(x_i,p_i)\}_{i=1}^{N}$, the privacy objective is defined as

\begin{equation}
    \mathcal{L}_{\mathrm{priv}}(\psi;\theta)
    =
    \frac{1}{N}\sum_{i=1}^{N}
    \ell_{\mathrm{priv}}\!\left(
    g_{\psi}\!\left(T_{\theta}(x_i)\right),\, p_i
    \right),
\end{equation}
where $\ell_{\mathrm{priv}}$ is the cross-entropy loss. For a fixed minimization configuration $T_{\theta}$, we train $g_{\psi}$ to convergence and use its prediction performance as a measure of privacy leakage. Lower privacy inference performance indicates stronger privacy preservation.


\subsection{Selection of Pareto-Optimal Setting}
\label{sec:model_selection}

For each data minimization configuration, we evaluate three metrics: i) anomaly detection, measured by AUC, and ii) privacy leakage, measured by cMAP and F1 score. A higher AUC indicates better performance in anomaly detection, while lower values of cMAP and F1 score signify stronger privacy preservation. Therefore, selecting an appropriate data minimization configuration is framed as a multi-objective optimization problem rather than a decision based solely on a single metric. The selection of optimal settings for data minimization is summarized in Algorithm~\ref{alg:model_selection}.

Let $\mathcal{S}=\{S_1,S_2,\dots,S_N\}$ denote the set of experimental settings. For each setting $S_i \in \mathcal{S}$, we obtain a triplet of measurements, given by the following equation.

\begin{equation}
    \mathbf{m}_i = \big(A_i, F_i, C_i\big),
\end{equation}
where $A_i$ denotes the AUC, $F_i$ and $C_i$ is the F1 score and cMAP, respectively. The selection objective is to:

\begin{equation}
    \max A_i, \qquad \min F_i, \qquad \min C_i.
\end{equation}
These objectives are conflicting in general: improving task performance may increase privacy leakage, while stronger privacy suppression may reduce task performance. Therefore, a principled selection strategy must account for the trade-off among all three criteria.

\subsubsection{Pareto optimal settings}

We first identify the set of Pareto-optimal settings. A setting $S_i$ is said to dominate another setting $S_j$ if

\begin{equation}
    A_i \geq A_j,\qquad F_i \leq F_j,\qquad C_i \leq C_j,
\end{equation}
and at least one of the above inequalities is strict. In other words, $S_i$ dominates $S_j$ if it performs at least as well in all objectives and is strictly better in at least one objective. Any setting that is dominated is considered suboptimal and can be excluded from further consideration. As a result, the Pareto set is defined as follows:

\begin{equation}
    \mathcal{P} = \left\{ S_i \in \mathcal{S} \; \middle| \; \nexists \, S_j \in \mathcal{S} \text{ such that } S_j \text{ dominates } S_i \right\}.
\end{equation}
This step removes clearly inferior settings and restricts model selection to the non-dominated frontier.

\subsubsection{Metric Normalization}

Since AUC, F1, and cMAP are measured on different scales, making direct aggregation inappropriate. To enable a meaningful joint analysis, we normalize all metrics to the range $[0,1]$, with higher values indicating better overall behavior. For AUC, which should be maximum, we use standard min-max normalization using following equation:

\begin{equation}
    \tilde{A}_i = \frac{A_i - \min_k A_k}{\max_k A_k - \min_k A_k}.
    \vspace{-1.5mm}
\end{equation}
For privacy metrics, where lower values are better, we reverse the scale after normalization and given by the following equation:

\begin{equation}
    \tilde{F}_i = \frac{\max_k F_k - F_i}{\max_k F_k - \min_k F_k},
\end{equation}

\begin{equation}
    \tilde{C}_i = \frac{\max_k C_k - C_i}{\max_k C_k - \min_k C_k}.
\end{equation}
After normalization, all three quantities $\tilde{A}_i$, $\tilde{F}_i$, and $\tilde{C}_i$ lie in $[0,1]$, where higher value is better.
\vspace{1mm}


\begin{algorithm}[h]
\caption{Selection of Pareto-optimal settings}
\label{alg:model_selection}
\small
\KwIn{Experimental settings $\mathcal{S}=\{S_1,\dots,S_N\}$ with metrics $(A_i,F_i,C_i)$}
\KwOut{Optimal operating point $s^*$}

\textbf{Step 1: Pareto filtering} \\

Construct the Pareto set
\begin{equation*}
    \mathcal{P} = \left\{ S_i \in \mathcal{S} \;\middle|\; \nexists\, s_j \in \mathcal{S}
    \text{ such that }
    A_j \geq A_i,
    F_j \leq F_i,
    C_j \leq C_i
    \right\}
\end{equation*}

where at least one inequality is strict for domination.

\textbf{Step 2: Metric normalization.} \\
For each $S_i \in \mathcal{P}$, compute

\begin{equation*}
    \tilde{A}_i=\frac{A_i-\min_k A_k}{\max_k A_k-\min_k A_k},
\end{equation*}

\begin{equation*}
    \tilde{F}_i=\frac{\max_k F_k-F_i}{\max_k F_k-\min_k F_k},
\end{equation*}

\begin{equation*}
    \tilde{C}_i=\frac{\max_k C_k-C_i}{\max_k C_k-\min_k C_k}.
\end{equation*}

\textbf{Step 3: Final operating point selection.} \\

\text{1. Ideal-point criterion}

\begin{equation*}
    D_i=\sqrt{(1-\tilde{A}_i)^2+(1-\tilde{F}_i)^2+(1-\tilde{C}_i)^2},
\end{equation*}

\begin{equation*}
    s^*=\arg\min_{s_i\in\mathcal{P}} D_i
\end{equation*}

\text{2. Weighted aggregation}
\begin{equation*}
    S_i=w_A\tilde{A}_i+w_F\tilde{F}_i+w_C\tilde{C}_i,
    \quad
    w_A+w_F+w_C=1,
\end{equation*}

\begin{equation*}
    s^*=\arg\max_{s_i\in\mathcal{P}} S_i
\end{equation*}

\text{3. Constrained-based selection}
\begin{equation*}
    s^*=\arg\max_{s_i\in\mathcal{S}} A_i
    \quad
    \text{s.t. }
    F_i\leq\tau_F,\;
    C_i\leq\tau_C.
\end{equation*}

\Return{$s^*$}
\end{algorithm}


\subsubsection{Optimal Setting Selection}

To identify the best optimal minimization configuration, we implement three selection strategies. i) Selection by distance to the ideal point, which ranks non-dominated configurations based on their distance to an ideal utility-privacy solution; ii) weighted aggregation-based Pareto selection, which combines normalized objectives into a single score using predefined utility-privacy weights; and iii) constraint-based selection, which filters configurations according to application-specific requirements and selects the final operating point from the feasible set.\\
\textbf{Selection by distance to the ideal point:} An optimal operating point can be selected by measuring the distance of each Pareto-optimal setting to the ideal point. In normalized space, the ideal point is

\begin{equation}
    \mathbf{u}^* = (1,1,1),
\end{equation}

which corresponds to maximal normalized AUC and minimal privacy leakage. For each setting $S_i$, we compute its Euclidean distance to the ideal point by the following equation:

\begin{equation}
    D_i = \sqrt{(1-\tilde{A}_i)^2 + (1-\tilde{F}_i)^2 + (1-\tilde{C}_i)^2 }.
\end{equation}

Therefore, the selected optimal setting is:

\begin{equation}
    s^* = \arg\min_{s_i \in \mathcal{P}} D_i.
    \label{distance_ideal_point}
\end{equation}

This criterion favors settings that achieve the best overall balance between anomaly detection and privacy preservation.\\
\textbf{Weighted aggregation based selection:} In this selection strategy, we aggregate the normalized metrics into a single scalar score using predefined weights and given by the following equation:

\begin{equation}
    S_i = w_A \tilde{A}_i + w_F \tilde{F}_i + w_C \tilde{C}_i,
\end{equation}

subject to

\begin{equation}
    w_A + w_F + w_C = 1, \qquad w_A,w_F,w_C \geq 0
\end{equation}

Therefore, the selected setting is:

\begin{equation}
    s^* = \arg\max_{s_i \in \mathcal{P}} S_i.
    \label{weighted_aggregation}
\end{equation}

This formulation is beneficial when the relative importance of utility tasks and privacy leakage is known in advance. For example, assigning a higher value to $w_A$ enhances anomaly detection performance, while larger values of $w_F$ and $w_C$ prioritize privacy preservation.\\
\textbf{iii) Constraint-based selection:} In this strategy, we consider constraint-based selection, where the data minimization configuration is chosen to satisfy predefined utility or privacy requirements:

\begin{equation}
    s^* = \arg\max_{s_i \in \mathcal{S}} A_i
\end{equation}

subject to

\begin{equation}
    F_i \leq \tau_F, \qquad C_i \leq \tau_C,
\end{equation}

where $\tau_F$ and $\tau_C$ are acceptable privacy thresholds. This strategy selects the highest-utility model among all settings that satisfy the required privacy constraints.


\section{Experiments}

\subsection{Dataset}
To train and evaluate the anomaly detection model, we used \textbf{UCSD Ped1}, \textbf{UCSD Ped2} \cite{5539872} and \textbf{CUHK Avenue} \cite{lu2013abnormal} datasets. The UCSD Ped1 dataset consists of 34 training and 36 testing videos, while UCSD Ped2 includes 16 training and 12 testing videos of varying lengths. The CUHK Avenue dataset contains 16 training and 21 testing videos. All datasets were captured using stationary cameras with fixed viewpoints. Training videos contain only normal events, whereas testing videos include both normal and abnormal activities, allowing for effective evaluation of anomaly detection performance.

The \textbf{PAHMDB} \cite{wu2020privacy} dataset is a subset of HMDB51 and comprises 515 videos used for both training and evaluation. It provides frame-level annotations for multiple privacy-sensitive attributes, including skin tone, relationships, face, nudity, and gender.

\subsection{Implementation Details}

We evaluate data minimization under a unified cross‑data training and evaluation protocol \cite{aslam2025balancing, dave2022spact, wu2020privacy} using the UCSD Pedestrian (Ped1 and Ped2), CUHK Avenue, and PAHMDB datasets. All models are trained for 50 epochs with the Adam \cite{kingma2014adam} optimizer with a learning rate of $1\times10^{-4}$. The training and evaluation batch sizes are 16 and 8, respectively. Each input sample consists of a 10‑frame video clip. For anomaly detection, we adopt a spatiotemporal autoencoder due to its simplicity and strong performance in video anomaly detection tasks \cite{chong2017abnormal}. A ResNet‑50 classifier is used as the private attribute detector \cite{he2016deep}.

Breadth‑based minimization is implemented using temporal sampling by increasing the stride used to create clip, while keeping the clip length fixed at 10 frames. We evaluate strides of $s = 5$ and $s = 10$, representing progressively stronger temporal reduction. Depth‑based minimization is implemented through spatial and appearance‑level transformations. For spatial minimization, inputs are downsample$\times$2 and downsample$\times$4. For appearance minimization, we consider masking, blur, and background removal. In the masking configuration, human regions are detected in each frame using YOLO11s‑segmentation \cite{ultralytics_github}, and all pixels within the detected person masks are suppressed to yield a simplified, binary‑like representation that preserves the surrounding scene. In the blur configuration, detected human regions are anonymized by applying Gaussian blur within the masks, obscuring identity‑related details while retaining coarse motion cues and spatial context. For background removal, background subtraction is used to suppress static content, retaining primarily the moving foreground. All experiments are conducted on NVIDIA V100 GPU using the PyTorch framework.

\subsection{Training and Evaluation Protocol}
To evaluate the performance of the proposed anomaly detection framework, we follow the standard cross-dataset training and evaluation protocol adopted in prior work \cite{dave2022spact, aslam2025balancing, wu2020privacy}. Under this protocol, the anomaly detector is trained and evaluated using anomaly detection datasets, whereas the privacy detector is trained and evaluated on a dedicated private-attribute dataset. Since private attribute annotations are not available for anomaly detection datasets, we adopt this cross-dataset strategy to ensure a fair and realistic evaluation, consistent with existing literature. Specifically, the anomaly detector is trained solely on anomaly data, while the privacy detector is trained on privacy-annotated data. We first train a baseline model without applying any data minimization. Subsequently, the same data minimization strategy applied to the anomaly data is also applied to the private dataset to enable a fair comparison of privacy leakage. Anomaly detection performance is evaluated using the Area Under the ROC Curve (AUC). Privacy leakage is assessed by measuring the performance of the target privacy model on the private-attribute test set using class-wise mean Average Precision (cMAP) and the F1-score.

\subsection{Results}

\begin{table*}[h]
\centering
\begin{tabular}{c|c|c|c|c|cc}
\hline
\multirow{2}{*}{Method} 
& \multirow{2}{*}{Setting} 
& UCSD Ped1 
& UCSD Ped2 
& CUHK Avenue 
& \multicolumn{2}{c}{PAHMDB} \\ 
& 
& AUC ($\uparrow$\%) 
& AUC ($\uparrow$\%) 
& AUC ($\uparrow$\%) 
& cMAP ($\downarrow$\%) 
& F1 ($\downarrow$\%) \\ \hline

Raw 
& -- 
& 79.10
& 92.49
& 83.01
& 70.20
& 0.396 \\ \hline

\multirow{2}{*}{Temporal sampling} 
& $s = 5$ 
& \textbf{72.98}
& \textbf{85.87}
& \textbf{77.06}
& 68.98
& 0.329\\ 
& $s = 10$ 
& 67.49
& 84.79
& 73.43
& 68.24
& 0.321\\ \hline

Downsample $\times 2$ 
& -- 
& 68.98
& 67.07
& 73.86
& 65.34
& 0.321\\ 

Downsample $\times 4$ 
& -- 
& 65.21
& 63.40
& 72.17
& 64.32
& 0.301\\ 

Masking 
& -- 
& 57.25
& 70.01
& 68.37
& 59.10
& 0.198 \\ 

Blur 
& -- 
& 72.15
& 81.28
& 75.60
& 62.35
& 0.211 \\ 

Background removal 
& -- 
& 52.18
& 58.27
& 50.62
& 68.23
& 0.259\\ \hline

\multirow{2}{*}{\begin{tabular}[c]{@{}c@{}}Temporal sampling +\\ Masking\end{tabular}} 
& $s = 5$ 
& 55.81
& 67.98
& 66.88
& 57.23
& 0.189\\ 
& $s = 10$ 
& 54.13
& 66.11
& 62.24
& \textbf{56.94}
& \textbf{0.184}\\ \hline

\multirow{2}{*}{\begin{tabular}[c]{@{}c@{}}Temporal sampling +\\ Blur\end{tabular}} 
& $\textcolor{blue}{s = 5}$ 
& \textcolor{blue}{71.04}
& \textcolor{blue}{84.25}
& \textcolor{blue}{73.93}
& \textcolor{blue}{61.35}
& \textcolor{blue}{0.201}\\ 
& $s = 10$ 
& 70.32
& 82.21
& 72.76
& 61.15
& 0.206\\ \hline

\multirow{2}{*}{\begin{tabular}[c]{@{}c@{}}Temporal sampling +\\ Background removal\end{tabular}} 
& $s = 5$ 
& 51.48
& 56.98
& 50.19
& 66.23
& 0.249\\ 
& $s = 10$ 
& 50.11
& 55.23
& 48.88
& 66.95
& 0.244\\ \hline

\end{tabular}
\caption{Performance under breadth- and depth-based data minimization. AUC is reported for video anomaly detection on UCSD Ped1, UCSD Ped2, and CUHK Avenue, while privacy metrics cMAP and F1 are reported on PAHMDB. Higher VAD performance and lower privacy performance are considered better and represented in \textbf{bold}. The best optimal trade-off is represented in \textcolor{blue}{blue} and $s$ denotes the stride used in temporal subsampling.}
\label{tab:minimization_results}
\end{table*}

\subsubsection{Comparison of Data Minimization Configurations}

Table~\ref{tab:minimization_results} highlights the trade-off between anomaly detection utility and privacy leakage across the evaluated data minimization settings. The raw-video setting achieves the best anomaly detection performance on UCSD Ped1, UCSD Ped2, and CUHK Avenue, but also produces the highest privacy leakage on PAHMDB, indicating that full visual and temporal information benefits utility while exposing the most privacy-sensitive cues.

Among breadth-based minimization methods, temporal sampling gradually reduces anomaly detection performance as the stride increases, while only providing limited privacy improvement. This suggests that temporal subsampling removes some redundant motion information, but does not sufficiently suppress privacy-relevant appearance.

For depth-based minimization, the effects vary by transformation type. Downsampling lowers AUC consistently with only moderate privacy gains. Masking provides the strongest privacy protection, substantially reducing cMAP and F1, but also causes a noticeable drop in anomaly detection performance. Blur offers a better balance, preserving relatively strong anomaly detection results while still significantly reducing privacy leakage. In contrast, background removal leads to the largest utility degradation without achieving the best privacy protection, indicating the importance of scene context for anomaly detection.

The combined settings further emphasize this trade-off. Temporal sampling with masking achieves the lowest privacy leakage, particularly at \(s=10\), but with a substantial loss in utility. Temporal sampling with blur provides a more balanced outcome: the \(s=5\) + blur setting maintains comparatively strong anomaly detection performance while still reducing privacy leakage considerably. The \(s=10\) + blur variant gives similar privacy benefits but slightly lower utility. Combinations with background removal remain less effective, as they operate in a low-utility regime without privacy advantages.

Overall, the results show that raw videos maximize utility but leak the most privacy, masking-based settings provide the strongest privacy protection at the cost of utility, and blur-based minimization offers the most favorable balance. Among all evaluated settings, temporal sampling with \(s=5\) combined with blur appears to be the most practical compromise.

\subsubsection{Analysis of Pareto-Optimal Setting}
\label{sec:Pareto_Optimal}

\begin{figure*}[h]
    \centering
    \includegraphics[width=0.945\linewidth]{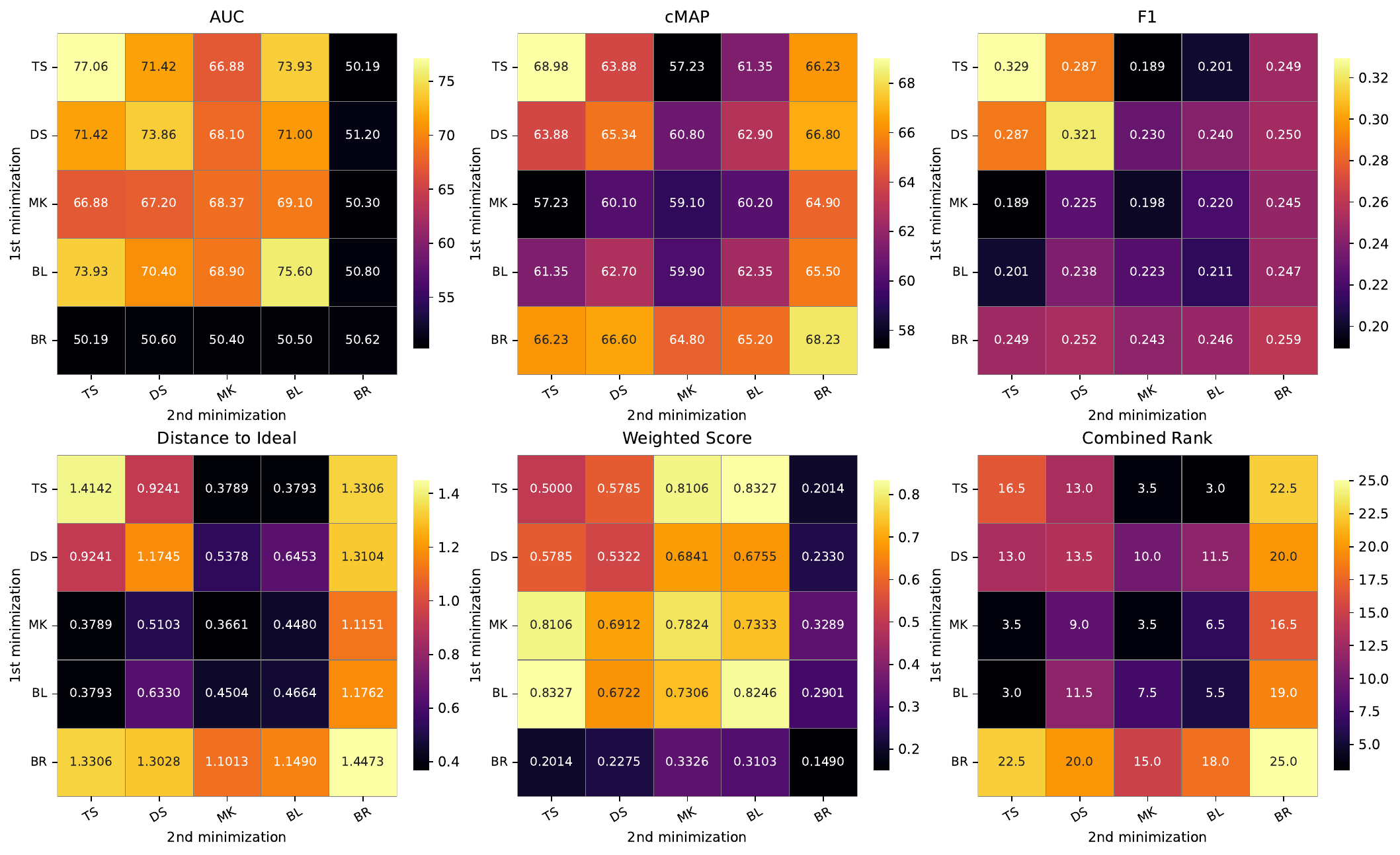}
    \caption{Heatmap visualization of raw performance metrics and derived selection criteria for all single and pairwise minimization settings. The top row shows AUC, cMAP, and F1, while the bottom row reports distance to the ideal point, weighted score, and combined rank. Diagonal entries denote single minimization settings, and off-diagonal entries denote pairwise settings, where the minimization on the vertical axis is applied first, and the minimization on the horizontal axis is applied second. TS: Temporal sampling; DS: Downsample; MK: Masking; BL: Blurring; BR: Background removal.}
    \label{fig:transformation_heatmap}
\end{figure*}

\begin{figure*}[htp]
    \centering
    \includegraphics[width=0.95\textwidth]{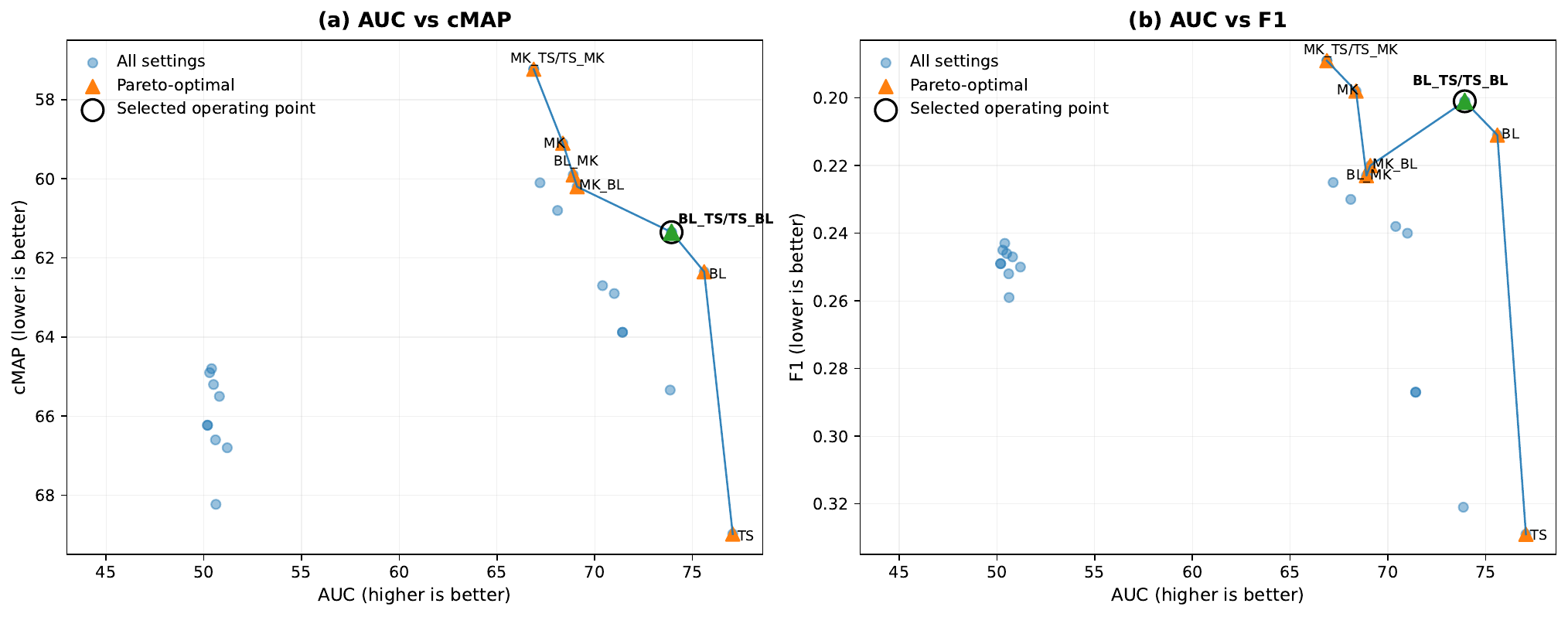}
    \caption{Projection of all candidate settings onto the AUC-cMAP and AUC-F1 planes. Higher AUC and lower cMAP/F1 are preferred. Highlighted markers denote settings that are Pareto-optimal in the full three-objective space of AUC, cMAP, and F1. The figure provides a compact visualization of the feasible utility-privacy trade-off surface and complements the quantitative ranking reported in Section~\ref{sec:Pareto_Optimal}.}
    \label{fig:pareto_projections}
\end{figure*}

Figure~\ref{fig:transformation_heatmap} summarizes the behavior of all candidate minimization settings under both raw evaluation metrics and derived model-selection criteria. The first row reports anomaly-detection utility on CUHK Avenue in terms of AUC, together with privacy leakage on PA-HMDB in terms of cMAP and F1. The second row reports the distance-to-ideal criterion from Eq.~\eqref{distance_ideal_point}, the weighted aggregation score from Eq.~\eqref{weighted_aggregation}, and a combined rank that consolidates both criteria into a single ordering. Specifically, for each setting $S_i$, we first compute its rank under distance to the ideal point, denoted by $r_i^{D}$, where a lower $D_i$ implies a better rank, and its rank under weighted score, denoted by $r_i^{W}$, where a higher weighted score implies a better rank. The final combined rank is then obtained by:

\begin{equation}
    r_i^{\text{comb}} = \frac{r_i^{D} + r_i^{W}}{2}.
\end{equation}
The lowest ranking score indicates the most optimal minimization settings.

Figure~\ref{fig:transformation_heatmap} yields three notable observations. First, masking-based minimization, either alone or in pairwise form, consistently yield lower privacy leakage and stronger selection scores than background-removal-based settings. Second, the pairwise settings TS$\rightarrow$BL and BL$\rightarrow$TS achieve the strongest weighted scores, whereas TS$\rightarrow$MK, MK, and MK$\rightarrow$TS remain highly competitive under the distance-based criterion. More importantly, under the combined-rank criterion, TS$\rightarrow$BL and BL$\rightarrow$TS emerge as the most favorable operating points, indicating the best overall balance between anomaly-detection utility and privacy preservation. This suggests that the combination of temporal sampling and blurring defines the most suitable utility-privacy trade-off among all evaluated settings and is therefore the preferred final operating point.

\subsubsection{Pareto Curve Analysis}

Figure~\ref{fig:pareto_projections} presents 2D Pareto curve analyses of the minimization settings, using AUC-cMAP and AUC-F1 as the corresponding utility-privacy objective pairs. In both subfigures, each point denotes one setting, and the highlighted triangular markers identify the non-dominated configurations in the respective 2D objective space. Specifically, a setting is Pareto-optimal in a given plane if no other setting simultaneously achieves higher AUC and lower privacy leakage in that plane. This representation makes the utility-privacy trade-off directly interpretable by explicitly revealing the feasible boundary beyond which no further improvement in one objective can be achieved without degrading the other.

Several trends are evident from the above Pareto curves. First, the competitive operating region is consistently concentrated in the upper-right portion of both curves, corresponding to high anomaly-detection utility with low privacy leakage. Second, masking- and blurring-based minimizations define the dominant trade-off boundary, whereas background-removal-based settings remain clearly inferior in both objective spaces. Third, the circled point corresponding to TS$\rightarrow$BL / BL$\rightarrow$TS lies on the Pareto frontier in both AUC-cMAP and AUC-F1 analyses and remains the most suitable operating point (sweet spot) when considered together with the ranking results in Section~\ref{sec:Pareto_Optimal}. This confirms that the combination of temporal sampling and blurring provides the most suitable overall balance between anomaly-detection utility and privacy preservation.


\section{Conclusion}

This paper presents a framework for data minimization in privacy-aware video anomaly detection, with a focus on the utility-privacy trade-off. We introduce a unified evaluation framework for breadth-based, depth-based, and combined minimization strategies using standard anomaly detection benchmarks together with a privacy-annotated dataset. Our results show that the choice of minimization strategy has a substantial impact on both anomaly detection utility and privacy leakage. To identify the most suitable data minimization configuration, we formulate the selection process as a multi-objective optimization problem. We analyze the resulting trade-off space through the Pareto frontier and apply three complementary selection strategies: distance-to-ideal-point, weighted aggregation, and constraint-based selection. Across all three strategies, the temporal sampling with stride s = 5 with blur is consistently identified as the optimal configuration and considered a sweet spot for the best trade-off. These findings demonstrate that privacy-by-design principles can be effectively incorporated into existing video anomaly detection pipelines, reducing privacy leakage while preserving useful anomaly detection performance.


{\small
\bibliographystyle{ieee_fullname}
\bibliography{references}
}

\end{document}